\documentclass{article}

\usepackage{arxiv}

\usepackage[utf8]{inputenc} 
\usepackage[T1]{fontenc}    
\usepackage{hyperref}       
\usepackage{url}            
\usepackage{booktabs}       
\usepackage{amsfonts}       
\usepackage{nicefrac}       
\usepackage{microtype}      
\usepackage{lipsum}
\usepackage{graphicx}
\graphicspath{ {./images/} }
\usepackage{algorithm}
\usepackage{algorithmic}
\usepackage{subcaption}

\usepackage{color,soul}
\usepackage{xcolor}

\title{Human and Multi-Agent collaboration in a human-MARL teaming framework
}
\author{
 Neda Navidi \\
  AI Redefined, AI-R Inc.\\
  400 McGill st., Montreal, Canada \\
  \texttt{neda@ai-r.com} \\
  \And
    François Chabot \\
    AI Redefined, AI-R Inc.\\
    400 McGill st., Montreal, Canada \\
    \texttt{Francois@ai-r.com} \\
  \And
    Sagar Kurandwad \\
    AI Redefined, AI-R Inc.\\
    400 McGill st., Montreal, Canada \\
    \texttt{Sagar@ai-r.com} \\
  \And
    Irv Lustigman \\
    AI Redefined, AI-R Inc.\\
    400 McGill st., Montreal, Canada \\
  \texttt{Irv@ai-r.com} \\
  \And
    Vincent Robert \\
    AI Redefined, AI-R Inc.\\
    400 McGill st., Montreal, Canada \\
  \texttt{Vincent@ai-r.com} \\
  \And
    Grégory Szriftgiser \\
    AI Redefined, AI-R Inc.\\
    400 McGill st., Montreal, Canada \\
    \texttt{Gregory@ai-r.com} \\
  \And
    Andrea Schuch \\
    AI Redefined, AI-R Inc.\\
    400 McGill st., Montreal, Canada \\
    \texttt{Andrea@ai-r.com} \\

}
\begin{document}
\maketitle
\begin{abstract}
Reinforcement learning provides effective results with agents learning from their observations, received rewards, and internal interactions between agents. This study proposes a new open-source MARL framework, called COGMENT, to efficiently leverage human and agent interactions as a source of learning. We demonstrate these innovations by using a designed real-time environment with unmanned aerial vehicles driven by RL agents, collaborating with a human. The results of this study show that the proposed collaborative paradigm and the open-source framework leads to significant reductions in both human effort and exploration costs. 
\end{abstract}
\section{Introduction}
An alternate approach in multi-agent systems entails the involvement of humans in the training process of RL agents known as human-in-the-loop (HITL). While studies have investigated human feedback impact and recommended its use to improve the sample efficiency of the single RL agent, as in imitation learning \cite{ross2011reduction} \cite{englert2013model} \cite{ross2010efficient}, behavioural cloning (BC) \cite{moore2019behavioural}, generative adversarial imitation learning (GAIL) \cite{ho2016generative}, we could not find any relevant study on human-AI teaming in multi-agent systems. A few studies have investigated the human-AI team concept in single agent systems, in which achieving high team performance depends on more than just the accuracy of the AI system. Since the human and the AI may have a different level of expertise, the highest team performance is often reached when they both know how and when to complement one another \cite{bansal2019beyond}.

The success of RL and multi-agent system (MAS) methods and algorithms in previous years highlights an essential need for scalable simulation platforms and RL frameworks. While different research groups have proposed RL frameworks which can manage the communication between the agents and the environment (OpenAI Gym \cite{brockman2016openai}, MuJoCo \cite{todorov2012mujoco}, Arcade Learning Environment \cite{bellemare2013arcade}, RLLib \cite{liang2017rllib}, Horizon \cite{gauci2018horizon}, and Dopamine \cite{castro2018dopamine}), the lack of a united human-MARL framework, usable for any type of environment design, and able to facilitate communication  between humans and the multi-agent system, prevents researchers from investigating and involving human expertise in multi-agent systems. 

The proposed solution is divided into two main components: leverage "human-MARL teaming framework", named COGMENT, to handle the collaboration between human and multiple RL-based agents in different environments. 

The remainder of the paper is organized as follows: Section 2 presents an overview of other related works. Section 3 describes the background methods of this study. Section 4 deals with the proposed methods and the contributions of this paper. Section 5 concludes this paper by summarizing the results achieved and provides suggestions for future work to improve the proposed methods.

\section{Related Works}
This section provides an overview of the key approaches this paper builds upon, namely, human-AI collaboration, multi-agent reinforcement learning, and reinforcement learning frameworks.

\subsection{MARL}
Generally, in MARL algorithms, the training mechanism is assigned to each agent separately.  This is referred to as decentralized or distributed learning. For example, independent Q learning can be employed to train each agent individually. A decentralized learning architecture can reduce the difficulty of the learning process and the complexity of the calculations \cite{arulkumaran2017brief}. \cite{tampuu2017multiagent} has considered the same concept in order to expand the framework of deep Q learning by dynamically adjusting the reward mode based on different goals and proposing a model in which multiple agents could cooperate and compete with each other.

In the face of a class of reasoning missions that require multiple agents to communicate with each other, the DQN model cannot usually learn effective strategies. \cite{foerster2016learning} has proposed a distributed deep loop Q network (DDRQN), which solves the problem of multi-agent communication and cooperation that can be observed in the state portion. Often, value-based RL algorithms are challenged by  inherently non-stationary environments, while policy gradient methods suffer from a variance that increases as the number of agents grows. In recent years, some new RL approaches have been proposed, including attention mechanism and Actor-Critic frameworks.

\cite{foerster2018counterfactual} proposed an actor-critic method called COMA policy gradients, where a centralized critic was used to estimate the Q-function, and decentralized actors used to optimize the agent policies. To address the challenges of multi-agent credit assignment, it uses a counterfactual baseline that marginalizes out a single agent’s action, while keeping the other agents’ actions fixed. 

\section{Background}
A Deep Q-learning is a class of off-policy Temporal difference ($TD$), consists of a multi-layer Deep Q-learning neural network ($DQN$) which states of an environment and action value may be its input and output, respectively. Q-learning $Q(s,\theta)$ updates $(\theta)$, the parameters  of $NN$ by observing next state $s^{'}$ after taking action $u^{a}\in U$ in state $s$ and receive an immediate reward $r$. Minimising the $TD$ error is considered to update the parameters of $(\theta)$, by:
\begin{equation}
    \Delta (\theta)= \sum_{i= 1}^{m}[((r + \gamma max_{u^{'}}Q(s^{'}, u^{'}; \theta^{'}))- Q(s, u; \theta))^{2}]
\end{equation}
where $m$ and $\theta^{'}$ present the sampling batch and parameters of target network, respectively.  This equation shows that a greedy policy is calculated by $max_{u^{'}}Q(s^{'}, u^{'}; \theta^{'}))- Q(s, u; \theta)$. The $max$ operator uses the same values to both select and evaluate an action, and it may lead to $overoptimistic$ value estimates \cite{van2016deep}.

To reduce the overestimation in DQN that can lead to inefficient learning policy, \cite{van2016deep} introduced double Q-learning $(DDQ-learning)$. DDQN can solve the problem by decomposing the $max$ operation of  $max_{u^{'}}Q(s^{'}, u^{'}$ into action selection and action evaluation, separately. Although not fully decoupled, the target network in the DQN architecture provides a natural candidate for the second value function, without having to introduce additional networks:
\begin{equation}
    \Delta (\theta)_{DDQN}= \sum_{i= 1}^{m}[((r + \gamma Q(s^{'}, argmax Q(s^{'},u^{'}; \theta);\theta^{'}))- Q(s, u; \theta))^{2}]
\end{equation}
 Also, the networks have essentially been trained on different samples from the memory bank of states and actions (the target network is “trained” on older samples than the primary network). Because of this, any bias due to environmental randomness should be smoothed out.

Multi-Agent Deep Deterministic Policy Gradient (MADDPG) algorithm is a  general-purpose multi-agent deep reinforcement learning algorithm based on policy gradient method \cite{lowe2017multi}. MADDPG does not assume a differentiable model of the environment dynamics or any particular structure on the communication method between agents. It can adopts the centralized training with decentralized execution to extend the Actor-Critic methods where the Critic is augmented with extra information about the policies of other agents, while the Actor only has access to local information. After training is completed, only the local actors are used at execution phase, acting in a decentralized manner.

Consider a mission scenario with $N$ agents with policies parameterized by $\theta =  (\theta_{1}, ..., \theta_{n})$, and let $\mu= (\mu_{1}, ..., \mu_{n})$ be the set of all agent policies. Then the gradient of the expected return $J(\theta_{i}) = \mathbb{E}[R_{i}]$  for agent $i$ can be written as:

\begin{equation}
     \nabla_{\theta_{i}}J(\mu_{i})= \mathbb{E}_{x, a\sim D} [\nabla_{\theta_{i}} \mu_{i} \nabla_{a_{i}} Q_{i}^{\mu}(X, a_{1}, ..., a_{N})|_{a_{i}=\mu_{i}(o_{i})}] \label{eq:MADDPG1}
\end{equation}
where, $Q_{i}^{\mu}(X, a_{1}, ..., a_{N})$ is a centralized action-value function that takes an input the actions of all agents, $a_{1}, ..., a_{N}$ in addition to some state information $X$, and outputs the Q-value for agent $i$. In the simplest case, $X$ could consist of the observations of all agents. The experience replay buffer $D$ contains the tuples $(X, X^{'}, a_{1}, ..., a_{N}, r_{1}, ..., r_{N})$ recording experiences of all agents. The parameters of policy network are updated using the policy gradient from Equation (\ref{eq:MADDPG1}) as the loss by an optimizer. 

\section{Methods}
\subsection{COGMENT}
\begin{figure} 
    \centering
    \includegraphics[width=150mm]{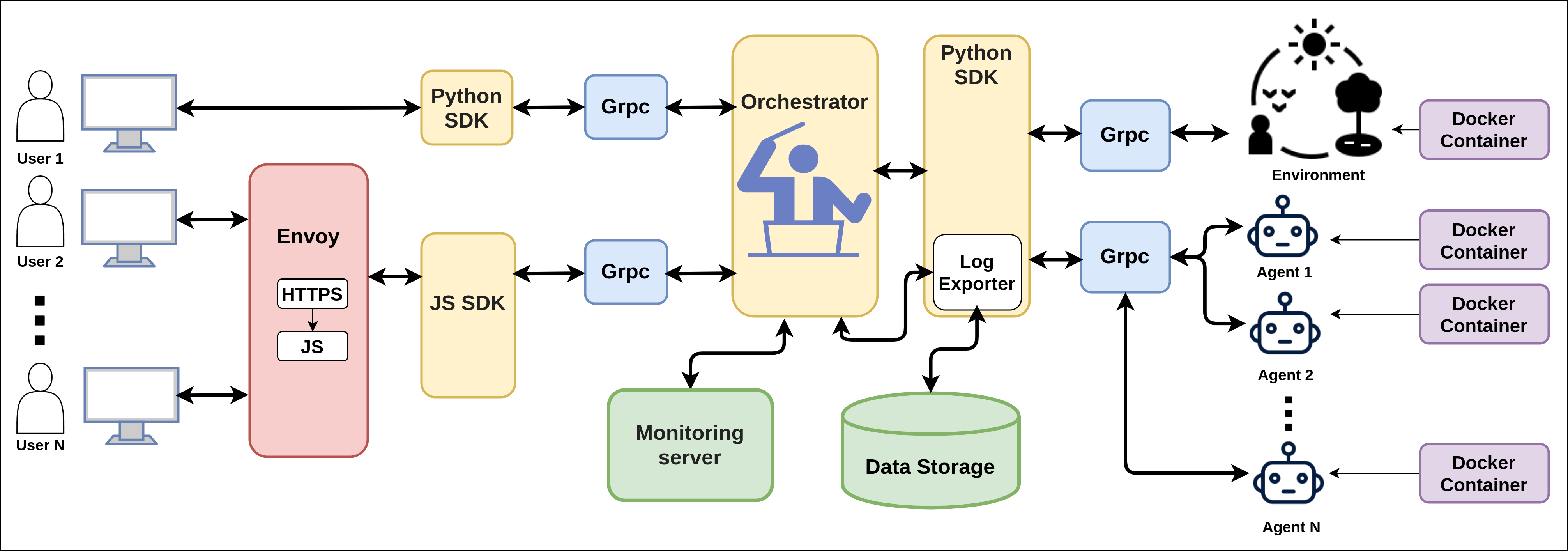}
    \caption{ Cogment framework}
    \label{fig:cogment}
\end{figure}

This study addresses the design, development, and demonstration of human-MARL teaming using a unique framework, called COGMENT ($See$ Figure \ref{fig:cogment}). The current version of COGMENT, developed by the AI Redefined team, was released as an open source framework to accelerate human-MARL teaming. COGMENT’s contribution can be segregated in four main parts: agents, user clients, and the orchestrator. Each part will be explained in the rest of this section.
 
Actors: In the COGMENT framework, actors can be defined as either human or agent. They  send actions, feedback, and/or recommendations to other actors; in the case of human users, through the shared environment by utilizing front-end clients (like a web client or a mobile app). COGMENT uses general-purpose Remote Procedure Calls (gRPC) to handle the communications between the different parts of the system. gRPC is a modern open source high performance RPC framework that can run in any environment. It can efficiently connect services to and across the “Orchestrator” — the application at the core of COGMENT — with pluggable support for load balancing, tracing, health checking and authentication. It is also applicable in the last mile of distributed computing to connect devices, mobile applications and browsers (usually the clients used by the users) to backend services. Cogment comes with SDKs (including APIs) to connect agents in python, and clients in either python or javascript to the environments. 

Agent and client: Heuristic and/or RL agents can communicate with clients (used by humans), other agents, and the environment through gRPC. Each agent can reside in its own separate docker-container. Using different docker-containers allows the deployment of agents on different machines in a distributed way, as well as individual modular configurations like the use of additional packages if they are needed by a specific type of agent.  including agents with heterogeneous capabilities. Since gRPC does not natively support javascript (JS), Envoy is used to connect to web-based front-end clients. The SDK also provides a way to store training data in a database ("log exporter"). COGMENT can support distributed users at a very large scale.

Orchestrator: The Orchestrator is the nexus of the whole system, as it manages all the communications, handles the combining of rewards from multiple sources, and generates offline datasets if needed. The orchestrator offers multiple options to address the shortcomings of the current RL framework, as well as active or passive training, multi or single agent systems. 
While the focus of this study is human-MARL teaming, COGMENT enables cooperative, assistive, and competitive interactions between humans and RL agents. COGMENT neither limits nor segregates the categories of interactions and allows them to be mixed up in a single use case. A monitoring server (which connects to the environment, the agents as well as many other components) can be used to evaluate the agent’s performance.

COGMENT also offers a command line interface called  “cogment-cli”, which provides a bootstrapping mechanism for setting up new projects easily.

\begin{figure} 
    \centering
    \includegraphics[width=100mm]{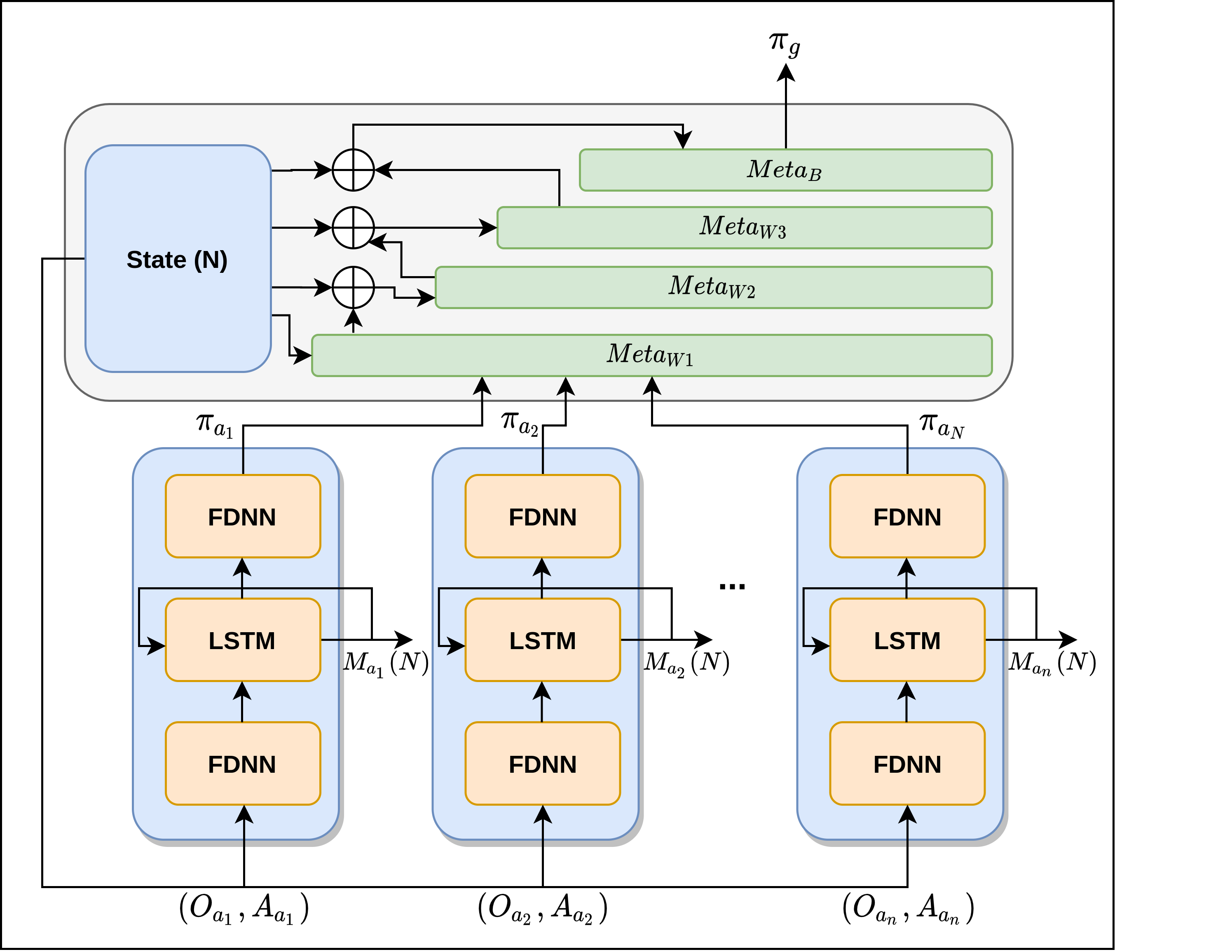}
    \caption{ Proposed Hybrid MARL}
\end{figure}

\subsection{Proposed Hybrid MARL}
This study propose a new MARL algorithm to take the benefit of both tightly-coupled and loosely-coupled MARL techniques. Inspired by MADDPG \cite{lowe2017multi}, this study proposes D3-MADDPG, by modifying the network of MADDPG, and replacing the Q-learning with Double Dueling Deep Q learning $(D3Q-learning)$. We aim to explicitly reduce the overestimation of Q-learning by a simple change in the target vale. Specifically, 
Also, the monotonic relationship in QMIX is guaranteed:
\begin{equation}
    \forall a\in A \frac{\partial Q_{MIX}(\tau,u)}{\partial Q_{a}}\geq 0
\end{equation}
where, $Q_{a}$ denotes a utility function, and $Q_{MIX}(\tau,u)$ is the mixing value, and presents ability to learn a joint action-value within linear complexity regarding the number of agents n. And,
\begin{equation}
    A^{\pi }(s,a)= Q_{MIX}^{\pi }(s,a)-V(s)
\end{equation}
where, $Q_{MIX}^{\pi }$ and $V(s)$, respectively, are:
\begin{equation}
    Q_{MIX}^{\pi }(s,a)= Q_{MIX}^{\pi }(s,a)-V(s)
\end{equation}

Recall that the Q value represents the value of choosing a specific action at a given state, and the V value represents the value of the given state regardless of the action taken. Then, intuitively, the Advantage value shows how advantageous selecting an action is relative to the others at the given state.

\section{Conclusion}
Although the current version of COGMENT simplifies the conduction of large scale experiments in HILT and/or multi-agent setups, through the current exercise, we aim to further the optimization of COGMENT’s communication channels, data storage/processing pipelines, and learning methodologies, in addition to providing a pool of continuous/discrete time simulated environments and pre-trained models of varying complexity to the research community.

\bibliographystyle{unsrt}  
\bibliography{mans}  

\begin{thebibliography}{10}

\bibitem{shalev2016safe}
Shai Shalev-Shwartz, Shaked Shammah, and Amnon Shashua.
\newblock Safe, multi-agent, reinforcement learning for autonomous driving.
\newblock {\em arXiv preprint arXiv:1610.03295}, 2016.

\bibitem{fridman2018deeptraffic}
Lex Fridman, Jack Terwilliger, and Benedikt Jenik.
\newblock Deeptraffic: Crowdsourced hyperparameter tuning of deep reinforcement
  learning systems for multi-agent dense traffic navigation.
\newblock {\em arXiv preprint arXiv:1801.02805}, 2018.

\bibitem{liang2019spectrum}
Le~Liang, Hao Ye, and Geoffrey~Ye Li.
\newblock Spectrum sharing in vehicular networks based on multi-agent
  reinforcement learning.
\newblock {\em IEEE Journal on Selected Areas in Communications},
  37(10):2282--2292, 2019.

\bibitem{cheng2019navigation}
WANG Cheng-bo, ZHANG Xin-yu, ZHANG Jia-wei, DING Zhi-guo, and AN~Lan-xuan.
\newblock Navigation behavioural decision-making of mass based on deep
  reinforcement learning and artificial potential field.
\newblock In {\em Journal of Physics: Conference Series}, volume 1357, page
  012026. IOP Publishing, 2019.

\bibitem{debard2019learning}
Quentin Debard, Jilles~Steeve Dibangoye, Stephane Canu, and Christian Wolf.
\newblock Learning 3d navigation protocols on touch interfaces with cooperative
  multi-agent reinforcement learning.
\newblock {\em arXiv preprint arXiv:1904.07802}, 2019.

\bibitem{chen2017decentralized}
Yu~Fan Chen, Miao Liu, Michael Everett, and Jonathan~P How.
\newblock Decentralized non-communicating multiagent collision avoidance with
  deep reinforcement learning.
\newblock In {\em 2017 IEEE international conference on robotics and automation
  (ICRA)}, pages 285--292. IEEE, 2017.

\bibitem{boutilier2018toward}
Craig Boutilier.
\newblock Toward user-centric recommender systems.
\newblock In {\em Proceedings of the 17th International Conference on
  Autonomous Agents and MultiAgent Systems}, pages 2--2. International
  Foundation for Autonomous Agents and Multiagent Systems, 2018.

\bibitem{okdinawati2017multi}
Liane Okdinawati, Togar~M Simatupang, and Yos Sunitiyoso.
\newblock Multi-agent reinforcement learning for collaborative transportation
  management (ctm).
\newblock In {\em Agent-Based Approaches in Economics and Social Complex
  Systems IX}, pages 123--136. Springer, 2017.

\bibitem{wang2016multi}
Hongbing Wang, Xiaojun Wang, Xingguo Hu, Xingzhi Zhang, and Mingzhu Gu.
\newblock A multi-agent reinforcement learning approach to dynamic service
  composition.
\newblock {\em Information Sciences}, 363:96--119, 2016.

\bibitem{tan1993multi}
Ming Tan.
\newblock Multi-agent reinforcement learning: Independent vs. cooperative
  agents.
\newblock In {\em Proceedings of the tenth international conference on machine
  learning}, pages 330--337, 1993.

\bibitem{bu2008comprehensive}
Lucian Bu, Robert Babu, Bart De~Schutter, et~al.
\newblock A comprehensive survey of multiagent reinforcement learning.
\newblock {\em IEEE Transactions on Systems, Man, and Cybernetics, Part C
  (Applications and Reviews)}, 38(2):156--172, 2008.

\bibitem{oliehoek2008optimal}
Frans~A Oliehoek, Matthijs~TJ Spaan, and Nikos Vlassis.
\newblock Optimal and approximate q-value functions for decentralized pomdps.
\newblock {\em Journal of Artificial Intelligence Research}, 32:289--353, 2008.

\bibitem{foerster2016learning}
Jakob Foerster, Ioannis~Alexandros Assael, Nando De~Freitas, and Shimon
  Whiteson.
\newblock Learning to communicate with deep multi-agent reinforcement learning.
\newblock In {\em Advances in neural information processing systems}, pages
  2137--2145, 2016.

\bibitem{foerster2018counterfactual}
Jakob~N Foerster, Gregory Farquhar, Triantafyllos Afouras, Nantas Nardelli, and
  Shimon Whiteson.
\newblock Counterfactual multi-agent policy gradients.
\newblock In {\em Thirty-second AAAI conference on artificial intelligence},
  2018.

\bibitem{lowe2017multi}
Ryan Lowe, Yi~I Wu, Aviv Tamar, Jean Harb, OpenAI~Pieter Abbeel, and Igor
  Mordatch.
\newblock Multi-agent actor-critic for mixed cooperative-competitive
  environments.
\newblock In {\em Advances in neural information processing systems}, pages
  6379--6390, 2017.

\bibitem{sunehag2017value}
Peter Sunehag, Guy Lever, Audrunas Gruslys, Wojciech~Marian Czarnecki, Vinicius
  Zambaldi, Max Jaderberg, Marc Lanctot, Nicolas Sonnerat, Joel~Z Leibo, Karl
  Tuyls, et~al.
\newblock Value-decomposition networks for cooperative multi-agent learning.
\newblock {\em arXiv preprint arXiv:1706.05296}, 2017.

\bibitem{sunehag2018value}
Peter Sunehag, Guy Lever, Audrunas Gruslys, Wojciech~Marian Czarnecki, Vinicius
  Zambaldi, Max Jaderberg, Marc Lanctot, Nicolas Sonnerat, Joel~Z Leibo, Karl
  Tuyls, et~al.
\newblock Value-decomposition networks for cooperative multi-agent learning
  based on team reward.
\newblock In {\em Proceedings of the 17th international conference on
  autonomous agents and multiagent systems}, pages 2085--2087. International
  Foundation for Autonomous Agents and Multiagent Systems, 2018.

\bibitem{rashid2018qmix}
Tabish Rashid, Mikayel Samvelyan, Christian~Schroeder De~Witt, Gregory
  Farquhar, Jakob Foerster, and Shimon Whiteson.
\newblock Qmix: monotonic value function factorisation for deep multi-agent
  reinforcement learning.
\newblock {\em arXiv preprint arXiv:1803.11485}, 2018.

\bibitem{son2019qtran}
Kyunghwan Son, Daewoo Kim, Wan~Ju Kang, David~Earl Hostallero, and Yung Yi.
\newblock Qtran: Learning to factorize with transformation for cooperative
  multi-agent reinforcement learning.
\newblock {\em arXiv preprint arXiv:1905.05408}, 2019.

\bibitem{mahajan2019maven}
Anuj Mahajan, Tabish Rashid, Mikayel Samvelyan, and Shimon Whiteson.
\newblock Maven: Multi-agent variational exploration.
\newblock In {\em Advances in Neural Information Processing Systems}, pages
  7611--7622, 2019.

\bibitem{van2016deep}
Hado Van~Hasselt, Arthur Guez, and David Silver.
\newblock Deep reinforcement learning with double q-learning.
\newblock In {\em Thirtieth AAAI conference on artificial intelligence}, 2016.

\bibitem{wang2015dueling}
Ziyu Wang, Tom Schaul, Matteo Hessel, Hado Van~Hasselt, Marc Lanctot, and Nando
  De~Freitas.
\newblock Dueling network architectures for deep reinforcement learning.
\newblock {\em arXiv preprint arXiv:1511.06581}, 2015.

\bibitem{ross2011reduction}
St{\'e}phane Ross, Geoffrey Gordon, and Drew Bagnell.
\newblock A reduction of imitation learning and structured prediction to
  no-regret online learning.
\newblock In {\em Proceedings of the fourteenth international conference on
  artificial intelligence and statistics}, pages 627--635, 2011.

\bibitem{englert2013model}
Peter Englert, Alexandros Paraschos, Jan Peters, and Marc~Peter Deisenroth.
\newblock Model-based imitation learning by probabilistic trajectory matching.
\newblock In {\em 2013 IEEE International Conference on Robotics and
  Automation}, pages 1922--1927. IEEE, 2013.

\bibitem{ross2010efficient}
St{\'e}phane Ross and Drew Bagnell.
\newblock Efficient reductions for imitation learning.
\newblock In {\em Proceedings of the thirteenth international conference on
  artificial intelligence and statistics}, pages 661--668, 2010.

\bibitem{moore2019behavioural}
Russell Moore, Andrew Caines, Andrew Rice, and Paula Buttery.
\newblock Behavioural cloning of teachers for automatic homework selection.
\newblock In {\em International Conference on Artificial Intelligence in
  Education}, pages 333--344. Springer, 2019.

\bibitem{vlachos2013investigation}
Andreas Vlachos.
\newblock An investigation of imitation learning algorithms for structured
  prediction.
\newblock In {\em European Workshop on Reinforcement Learning}, pages 143--154,
  2013.

\bibitem{ho2016generative}
Jonathan Ho and Stefano Ermon.
\newblock Generative adversarial imitation learning.
\newblock In {\em Advances in neural information processing systems}, pages
  4565--4573, 2016.

\bibitem{bansal2019beyond}
Gagan Bansal, Besmira Nushi, Ece Kamar, Walter~S Lasecki, Daniel~S Weld, and
  Eric Horvitz.
\newblock Beyond accuracy: The role of mental models in human-ai team
  performance.
\newblock In {\em Proceedings of the AAAI Conference on Human Computation and
  Crowdsourcing}, volume~7, pages 2--11, 2019.

\bibitem{brockman2016openai}
Greg Brockman, Vicki Cheung, Ludwig Pettersson, Jonas Schneider, John Schulman,
  Jie Tang, and Wojciech Zaremba.
\newblock Openai gym.
\newblock {\em arXiv preprint arXiv:1606.01540}, 2016.

\bibitem{todorov2012mujoco}
Emanuel Todorov, Tom Erez, and Yuval Tassa.
\newblock Mujoco: A physics engine for model-based control.
\newblock In {\em 2012 IEEE/RSJ International Conference on Intelligent Robots
  and Systems}, pages 5026--5033. IEEE, 2012.

\bibitem{bellemare2013arcade}
Marc~G Bellemare, Yavar Naddaf, Joel Veness, and Michael Bowling.
\newblock The arcade learning environment: An evaluation platform for general
  agents.
\newblock {\em Journal of Artificial Intelligence Research}, 47:253--279, 2013.

\bibitem{liang2017rllib}
Eric Liang, Richard Liaw, Philipp Moritz, Robert Nishihara, Roy Fox, Ken
  Goldberg, Joseph~E Gonzalez, Michael~I Jordan, and Ion Stoica.
\newblock Rllib: Abstractions for distributed reinforcement learning.
\newblock {\em arXiv preprint arXiv:1712.09381}, 2017.

\bibitem{gauci2018horizon}
Jason Gauci, Edoardo Conti, Yitao Liang, Kittipat Virochsiri, Yuchen He,
  Zachary Kaden, Vivek Narayanan, Xiaohui Ye, Zhengxing Chen, and Scott
  Fujimoto.
\newblock Horizon: Facebook's open source applied reinforcement learning
  platform.
\newblock {\em arXiv preprint arXiv:1811.00260}, 2018.

\bibitem{castro2018dopamine}
Pablo~Samuel Castro, Subhodeep Moitra, Carles Gelada, Saurabh Kumar, and Marc~G
  Bellemare.
\newblock Dopamine: A research framework for deep reinforcement learning.
\newblock {\em arXiv preprint arXiv:1812.06110}, 2018.

\bibitem{wang2018transferring}
Guo-fang Wang, Zhou Fang, Ping Li, and Bo~Li.
\newblock Transferring knowledge from human-demonstration trajectories to
  reinforcement learning.
\newblock {\em Transactions of the Institute of Measurement and Control},
  40(1):94--101, 2018.

\bibitem{ng1999policy}
Andrew~Y Ng, Daishi Harada, and Stuart Russell.
\newblock Policy invariance under reward transformations: Theory and
  application to reward shaping.
\newblock In {\em ICML}, volume~99, pages 278--287, 1999.

\bibitem{attia2018global}
Alexandre Attia and Sharone Dayan.
\newblock Global overview of imitation learning.
\newblock {\em arXiv preprint arXiv:1801.06503}, 2018.

\bibitem{zhang2016query}
Jiakai Zhang and Kyunghyun Cho.
\newblock Query-efficient imitation learning for end-to-end autonomous driving.
\newblock {\em arXiv preprint arXiv:1605.06450}, 2016.

\bibitem{bansal2019updates}
Gagan Bansal, Besmira Nushi, Ece Kamar, Daniel~S Weld, Walter~S Lasecki, and
  Eric Horvitz.
\newblock Updates in human-ai teams: Understanding and addressing the
  performance/compatibility tradeoff.
\newblock In {\em Proceedings of the AAAI Conference on Artificial
  Intelligence}, volume~33, pages 2429--2437, 2019.

\bibitem{arulkumaran2017brief}
Kai Arulkumaran, Marc~Peter Deisenroth, Miles Brundage, and Anil~Anthony
  Bharath.
\newblock A brief survey of deep reinforcement learning.
\newblock {\em arXiv preprint arXiv:1708.05866}, 2017.

\bibitem{tampuu2017multiagent}
Ardi Tampuu, Tambet Matiisen, Dorian Kodelja, Ilya Kuzovkin, Kristjan Korjus,
  Juhan Aru, Jaan Aru, and Raul Vicente.
\newblock Multiagent cooperation and competition with deep reinforcement
  learning.
\newblock {\em PloS one}, 12(4), 2017.

\bibitem{choi2017multi}
Jinyoung Choi, Beom-Jin Lee, and Byoung-Tak Zhang.
\newblock Multi-focus attention network for efficient deep reinforcement
  learning.
\newblock In {\em Workshops at the Thirty-First AAAI Conference on Artificial
  Intelligence}, 2017.

\bibitem{jiang2018learning}
Jiechuan Jiang and Zongqing Lu.
\newblock Learning attentional communication for multi-agent cooperation.
\newblock In {\em Advances in neural information processing systems}, pages
  7254--7264, 2018.

\bibitem{goldman2004decentralized}
Claudia~V Goldman and Shlomo Zilberstein.
\newblock Decentralized control of cooperative systems: Categorization and
  complexity analysis.
\newblock {\em Journal of artificial intelligence research}, 22:143--174, 2004.

\bibitem{gupta2017cooperative}
Jayesh~K Gupta, Maxim Egorov, and Mykel Kochenderfer.
\newblock Cooperative multi-agent control using deep reinforcement learning.
\newblock In {\em International Conference on Autonomous Agents and Multiagent
  Systems}, pages 66--83. Springer, 2017.

\end{thebibliography}

\end{document}